\documentclass{article} 
\usepackage[preprint]{colm2026_conference}

\usepackage{microtype}
\usepackage{hyperref}
\usepackage{url}
\usepackage{booktabs}
\usepackage{amsmath}
\usepackage{amssymb}
\usepackage{graphicx}
\usepackage{algorithm}
\usepackage{algorithmic}
\usepackage{multirow}
\usepackage{tikz}
\usetikzlibrary{arrows.meta,decorations.pathreplacing,calc,positioning}

\usepackage{lineno}

\definecolor{darkblue}{rgb}{0, 0, 0.5}
\hypersetup{colorlinks=true, citecolor=darkblue, linkcolor=darkblue, urlcolor=darkblue, unicode=true}

\newcommand{\method}{EqLen}

\title{Rethinking the Comparison Unit in Sequence-Level Reinforcement Learning: An Equal-Length Paired Training Framework from Loss Correction to Sample Construction}

\author{Fei Ding\thanks{Corresponding author: \texttt{dignfei@gmail.com}}, Yongkang Zhang, Runhao Liu\\
Alibaba Group\\
\And
Yuhao Liao, Zijian Zeng, Huiming Yang, Sibo Wang, Linglin Liao\\
Tsinghua University
}

\begin{document}

    \ifcolmsubmission
    \linenumbers
    \fi

    \maketitle

    \begin{abstract}
        This paper investigates the length problem in sequence-level relative reinforcement learning. We observe that, although existing methods partially alleviate length-related phenomena, a more fundamental issue remains insufficiently characterized: the comparison units used during training lack inherent comparability. Building on this observation, we propose a new perspective: the length problem should not be viewed merely as a loss-scaling or normalization bias, but rather as a \emph{comparison unit construction} problem. We further establish a sample-construction-based training framework that, instead of applying post-hoc corrections to unequal-length responses, proactively constructs equal-length, alignable, and comparable training segments during generation. Within this framework, we propose \method{}, a concrete method applicable to group-relative comparison algorithms such as GRPO, GSPO, and RLOO. Through dual-track synchronous generation, prefix inheritance, and segment masking, \method{} efficiently collects effective equal-length training segments and enables stable optimization.
    \end{abstract}

    \section{Introduction}

    Sequence-level relative reinforcement learning has become a key technique for enhancing the complex reasoning capabilities of large language models (LLMs). Among these methods, group-relative comparison-based optimization approaches have attracted widespread attention due to their training stability, implementation simplicity, and reduced dependence on reward models. These methods typically assume that multiple candidate responses generated for the same question can be directly compared and jointly optimized based on their relative quality. However, a critical issue remains insufficiently characterized: whether these candidate responses inherently constitute directly comparable training units. In language model generation, different responses often exhibit substantial variation in length, termination position, and reasoning depth. Consequently, the optimization process compares not only answer quality but also conflates structural misalignment and scaling distortion introduced by length differences. In other words, the root cause of existing length problems may lie not only in how the objective function is corrected, but in whether the training comparison units themselves are comparable.

    Building on this observation, we propose a new perspective: the length problem in sequence-level relative reinforcement learning should fundamentally be understood as a \textbf{comparison unit construction problem}, rather than merely a loss function correction problem. Existing work primarily mitigates length-related effects from the loss function side through reweighting, normalization, or length-scaling corrections. While these methods have achieved empirical progress, they still operate on direct comparisons of naturally unequal-length responses and therefore do not change the fact that comparison units are structurally misaligned. We argue that, rather than applying post-hoc compensation on incomparable samples, it is more effective to proactively construct more alignable and comparable training segments during generation. The length problem is thus reformulated as: how to design a training mechanism such that optimization is genuinely built upon structurally fair comparison units.

    Based on this perspective, we further establish a sample-construction-based training framework whose core idea is: instead of directly comparing naturally unequal-length complete responses, we proactively construct equal-length, alignable, and comparable training segments during the generation process. This framework shifts the treatment of the length problem from passive correction on the loss side to active construction on the sample side, thereby making the definition of training signals clearer and the comparison relationships more consistent. Within this framework, we propose the concrete method \method{}. This method collects candidate trajectories through a dual-track synchronous generation mechanism. When one trajectory terminates early, the effective prefix of the other trajectory is used to continue constructing new comparison branches, and multiple effective equal-length training segments are collected at length-aligned positions. Furthermore, we introduce a segment masking mechanism that excludes inherited prefixes and invalid suffixes after pair expiration from gradient updates, so that policy optimization acts only on newly generated suffixes that genuinely participate in comparison. Through this approach, \method{} translates the new perspective of ``comparison unit comparability'' into an executable training mechanism, significantly improving training signal utilization efficiency while mitigating length-induced optimization distortion.

    We propose a fundamentally different approach: \textbf{rather than patching the loss function on unequal-length trajectories, we generate equal-length trajectories from the source}. Our equal-length trajectory generation method (\method{}) modifies the inference framework itself: (1)~\textbf{Synchronous dual-track generation}: groups are split into subgroups of size 2, with two trajectories decoded synchronously; (2)~\textbf{Prefix inheritance}: when one trajectory terminates, the existing tokens of the other serve as a prefix to initialize a new trajectory for continued generation, and this process repeats; (3)~\textbf{Equal-length pairing}: each ``convergence point'' naturally produces an equal-length trajectory pair that is directly used for training.

\begin{figure}[t]
        \centering
        \begin{tikzpicture}[
            >=Stealth,
            lbl/.style={font=\scriptsize, inner sep=1pt},
            brace/.style={decorate, decoration={brace, amplitude=3pt, raise=1.5pt}},
        ]
        \node[font=\small\bfseries, anchor=west] at (0, 3.6) {(a) Standard GRPO ($G\!=\!4$)};

        \def\xstart{0.5}

        \fill[teal!65, rounded corners=1.5pt]  (\xstart, 3.05) rectangle (1.7, 3.35);
        \fill[red!45, rounded corners=1.5pt]   (\xstart, 2.35) rectangle (4.5, 2.65);
        \fill[teal!65, rounded corners=1.5pt]  (\xstart, 1.65) rectangle (2.9, 1.95);
        \fill[red!45, rounded corners=1.5pt]   (\xstart, 0.95) rectangle (6.0, 1.25);

        \node[lbl, anchor=east] at (\xstart-0.08, 3.2) {$o_1$};
        \node[lbl, anchor=east] at (\xstart-0.08, 2.5) {$o_2$};
        \node[lbl, anchor=east] at (\xstart-0.08, 1.8) {$o_3$};
        \node[lbl, anchor=east] at (\xstart-0.08, 1.1) {$o_4$};

        \node[lbl, anchor=west] at (1.8, 3.2)  {$L\!=\!50,\ r\!=\!1$};
        \node[lbl, anchor=west] at (4.6, 2.5)  {$L\!=\!180,\ r\!=\!0$};
        \node[lbl, anchor=west] at (3.0, 1.8)  {$L\!=\!100,\ r\!=\!1$};
        \node[lbl, anchor=west] at (6.1, 1.1)  {$L\!=\!250,\ r\!=\!0$};

        \draw[brace, red!60!black] (7.6, 3.4) -- (7.6, 0.9)
            node[midway, right=5pt, font=\scriptsize, align=left, text=red!60!black] {Unequal lengths\\[-1pt]Biased gradients};

        \draw[gray!35, dashed, line width=0.5pt] (-0.1, 0.45) -- (8.5, 0.45);

        \node[font=\small\bfseries, anchor=west] at (0, 0.0) {(b) \method{} ($G\!=\!2$, synchronous generation)};

        \def\ya{-0.75}  
        \def\yb{-1.5}   

        \node[lbl, anchor=east, font=\scriptsize\bfseries] at (\xstart-0.08, \ya+0.15) {$A$};
        \node[lbl, anchor=east, font=\scriptsize\bfseries] at (\xstart-0.08, \yb+0.15) {$B$};

        \fill[blue!45, rounded corners=1.5pt] (\xstart, \ya) rectangle (2.8, \ya+0.3);
        \node[font=\tiny\bfseries, white] at (1.65, \ya+0.15) {$o_{A_1}$};
        \draw[red!70!black, line width=1pt] (2.8, \ya-0.02) -- (2.8, \ya+0.32);
        \node[font=\tiny, red!70!black, anchor=south west] at (2.82, \ya+0.28) {\textsf{EOS}};

        \fill[orange!55, rounded corners=1.5pt] (\xstart, \yb) rectangle (2.8, \yb+0.3);
        \node[font=\tiny\bfseries, white] at (1.65, \yb+0.15) {$o_{B_1}$};
        \fill[orange!25, rounded corners=1.5pt] (2.8, \yb) rectangle (5.0, \yb+0.3);
        \node[font=\tiny, orange!60!black] at (3.9, \yb+0.15) {$B$ cont.};

        \draw[brace, teal!60!black] (2.85, \ya+0.3) -- (2.85, \yb)
            node[midway, right=4pt, font=\tiny\bfseries, teal!60!black] {Pair 1: equal length $T_1$};

        \draw[dashed, gray!50, line width=0.5pt] (2.8, \ya-0.02) -- (2.8, \yb-0.02);

        \draw[->, gray!55, line width=0.8pt, rounded corners=4pt]
            (5.0, \yb+0.15) -- (5.5, \yb+0.15) -- (5.5, \yb-0.6) -- (\xstart+0.15, \yb-0.6) -- (\xstart+0.15, \yb-0.85);
        \node[font=\tiny, gray!185, anchor=south] at (2.8, \yb-0.7) {Prefix inheritance};

        \def\yap{-2.8}  
        \def\ybp{-3.5}  

        \node[lbl, anchor=east, font=\scriptsize\bfseries] at (\xstart-0.08, \yap+0.15) {$A'$};
        \node[lbl, anchor=east, font=\scriptsize\bfseries] at (\xstart-0.08, \ybp+0.15) {$B$};

        \fill[gray!18, rounded corners=1.5pt] (\xstart, \yap) rectangle (2.8, \yap+0.3);
        \draw[gray!40, line width=0.4pt] (\xstart, \yap) rectangle (2.8, \yap+0.3);
        \node[font=\tiny, gray!50!black] at (1.55, \yap+0.15) {Prefix};
        \fill[blue!45, rounded corners=1.5pt] (2.8, \yap) rectangle (5.2, \yap+0.3);
        \node[font=\tiny\bfseries, white] at (4.0, \yap+0.15) {$o_{A_2}$};

        \fill[gray!18, rounded corners=1.5pt] (\xstart, \ybp) rectangle (2.8, \ybp+0.3);
        \draw[gray!40, line width=0.4pt] (\xstart, \ybp) rectangle (2.8, \ybp+0.3);
        \node[font=\tiny, gray!50!black] at (1.55, \ybp+0.15) {Prefix};
        \fill[orange!55, rounded corners=1.5pt] (2.8, \ybp) rectangle (5.2, \ybp+0.3);
        \node[font=\tiny\bfseries, white] at (4.0, \ybp+0.15) {$o_{B_2}$};
        \draw[red!70!black, line width=1pt] (5.2, \ybp-0.02) -- (5.2, \ybp+0.32);
        \node[font=\tiny, red!70!black, anchor=south west] at (5.22, \ybp+0.28) {\textsf{EOS}};

        \draw[brace, teal!60!black] (6.15, \yap+0.3) -- (6.15, \ybp)
            node[midway, right=4pt, font=\tiny\bfseries, teal!60!black] {Pair 2: equal length $T_2$};

        \draw[dashed, gray!50, line width=0.5pt] (5.2, \yap-0.02) -- (5.2, \ybp-0.02);

        \node[font=\normalsize, gray!50] at (5.8, \ybp+0.15) {$\cdots$};

        \node[draw=teal!50!black, rounded corners=3pt, fill=teal!5,
              font=\scriptsize, align=center, text=teal!60!black,
              inner sep=4pt, anchor=west] at (6.2, \yb-0.3)
            {All pairs\\[-1pt]equal-length, unbiased};

        \end{tikzpicture}
        \caption{Overview of the \method{} mechanism. (a)~Standard GRPO samples 4 trajectories of vastly different lengths ($L \in [50, 250]$), resulting in biased gradient computation. (b)~\method{} synchronously generates two trajectories, harvesting equal-length pairs at each EOS convergence point and inheriting the prefix for continued generation, cyclically producing multiple equal-length trajectory pairs. Gray regions denote inherited prefixes (excluded from gradient computation).}\label{fig:overview}
    \end{figure}
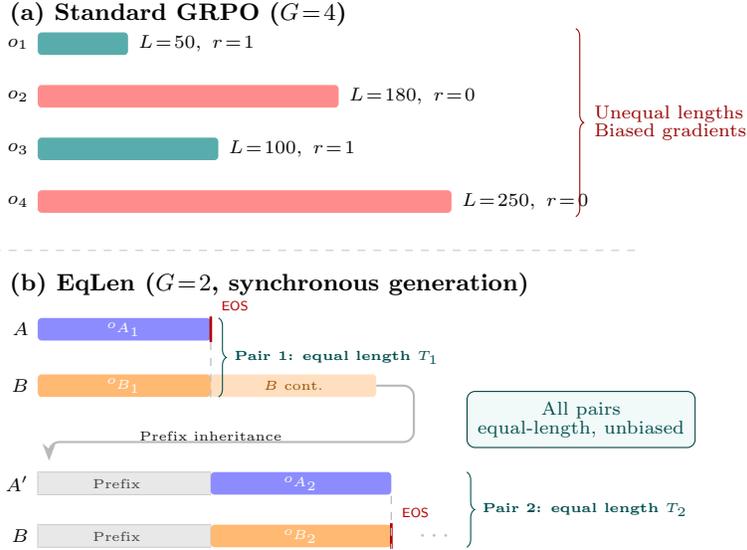

    Figure~\ref{fig:overview} provides an intuitive comparison between \method{} and standard GRPO.

    \paragraph{Our contributions are as follows:}
    \begin{itemize}
        \item We identify a critical yet insufficiently characterized problem: the length phenomena in existing sequence-level relative reinforcement learning originate not merely from loss scaling or normalization bias, but from the fundamental \textbf{lack of comparability} of the training comparison units themselves.

        \item We reveal two previously overlooked failure modes caused by length inconsistency in GRPO---\textbf{entropy collapse} (correct long-reasoning patterns are erroneously penalized due to trajectory truncation) and \textbf{wasted updates / learning tax} (shared segments oscillate between positive and negative gradients, squandering computational resources)---harms that go beyond the conventional understanding of ``favoring longer/shorter responses.''

        \item We propose a new problem perspective and, based on it, establish a sample-construction-based training framework: by proactively constructing equal-length, alignable, and comparable training segments during generation, we replace post-hoc loss corrections on unequal-length responses. Since this framework modifies the inference generation process rather than the loss function, it can be orthogonally combined with algorithm-level improvements such as Dr.~GRPO and DAPO.

        \item We propose the concrete method \method{} and validate its effectiveness on two architectures, Qwen3-32B and Qwen3-Next: it surpasses the strongest baseline by 5.4 percentage points on AIME25 and by 5.8 percentage points on HMMT25, achieves approximately 6$\times$ generation efficiency improvement (Section~\ref{sec:efficiency_exp}), and reaches 86.4\% when combined with DAPO.
    \end{itemize}

    \section{Related Work}

    \subsection{Reinforcement Learning for Large Language Models}

    Reinforcement learning from human feedback (RLHF)~\citep{NEURIPS2022_b1efde53} has become the standard paradigm for aligning LLMs, with PPO~\citep{schulman2017proximalpolicyoptimizationalgorithms} as its canonical implementation. GRPO~\citep{shao2024deepseekmathpushinglimitsmathematical} replaces the value function with group-relative rewards, substantially reducing training overhead, and has driven breakthrough emergent reasoning capabilities in DeepSeek-R1~\citep{Guo_2025}. However, GRPO's group-relative comparison mechanism makes it particularly sensitive to trajectory length differences---the core problem addressed in this paper.

    \subsection{Length Bias in Reinforcement Learning}

    Length bias has been widely studied in RLHF. \citet{singhal2024a} found that a reward function based solely on length can reproduce most of the downstream improvements of RLHF, revealing the severity of length bias. This problem is especially pronounced in GRPO, where length differences among trajectories within a group directly affect gradient computation.

    \textbf{Algorithm-level corrections.} Dr.~GRPO~\citep{liu2025understanding} identifies two specific sources of bias in GRPO---normalization by sequence length causes long incorrect responses to be insufficiently penalized, and normalization by standard deviation causes questions with extreme reward distributions to be overweighted---and proposes removing both. DAPO~\citep{yu2026dapo} introduces token-level policy gradient loss and an overlong reward shaping mechanism that penalizes responses exceeding a preset length threshold. GSPO~\citep{zheng2025groupsequencepolicyoptimization} argues theoretically that applying token-level importance sampling in GRPO is ill-posed because rewards are sequence-level, and therefore lifts clipping, rewards, and optimization entirely to the sequence level. However, LUSPO~\citep{liu2026lengthunbiasedsequencepolicyoptimization} finds that GSPO further amplifies length bias and corrects this by multiplying each sequence's loss by its length. P-GSPO~\citep{hu2025pgspo} introduces a power-law normalization parameter to control how length scales policy updates. DLER~\citep{liu2025dlerdoinglengthpenalty} revisits simple truncation penalties, finding that accuracy drops stem from insufficient optimization rather than the penalty form itself, and proposes batch-level reward normalization and difficulty-aware truncation. LSPO~\citep{chen2025lspolengthawaredynamicsampling} approaches the problem from a data selection perspective, dynamically selecting training data based on average response length. GR3~\citep{li2026tacklinglengthinflationtradeoffs} proposes a multiplicative reward scaling paradigm, arguing that additive penalties create compensatory optimization shortcuts.

    \textbf{Reward model-level corrections.} Another line of work addresses length bias in the reward model itself. \citet{bu-etal-2025-beyond} adaptively adjusts length influence based on query context; FiMi-RM~\citep{anonymous2026bias} reveals multi-stage nonlinear patterns of bias and fits debiasing models; \citet{Kim_Oh_Lee_2026} uses causal inference for counterfactual data augmentation; Rc-BT~\citep{cai2026disentangling} decouples semantic preference from length requirements; PAR~\citep{fu2025reward} replaces raw rewards with latent preferences. These methods operate solely on the reward model and do not address length bias at the RL optimization level.

    \textbf{Other methods addressing length-related pathologies.} Recent work tackles length-related training stability from different mechanistic angles. VESPO~\citep{shen2026vespovariationalsequencelevelsoft} models the reshaping of sequence-level importance sampling weights as a variational measure-change problem, deriving a closed-form soft suppression kernel $\phi(W) = W^{c_1} \cdot e^{c_2(1-W)}$ to replace PPO's hard clipping, avoiding the length bias introduced by $1/T$ normalization; however, it remains fundamentally a loss-function-level correction---the importance weights of long and short trajectories are inherently coupled with length, and the exponential kernel only smooths the clipping boundary without eliminating the fundamental influence of length on gradient magnitudes. T2T~\citep{lin2026boostingllmreasoninghumaninspired} proposes capability-aware two-stage length shaping: encouraging longer reasoning to expand the search space when the model answers incorrectly (thickening), and imposing length penalties to encourage conciseness when it answers correctly (thinning); this method's length regulation depends on the correctness of reward signals and may fail under noisy rewards, and it still builds upon comparisons of unequal-length trajectories. TreeRPO~\citep{yang2025treerpotreerelativepolicy} achieves PRM-free step-level credit assignment through tree-structured sampling: at each reasoning step, it branches $N$ candidates to build an $N$-ary tree, evaluates leaf nodes, and propagates expected rewards bottom-up; however, its computational cost is $O(N^D)$ ($D$ being the tree depth), far exceeding \method{}'s dual-track generation, and it has only been validated on a 1.5B model with unknown scalability. All the above methods alleviate length-related issues from various angles, but none guarantees equal length at the generation mechanism level---this is precisely the unique contribution of \method{}.

    \textbf{Common limitations of existing methods.} Table~\ref{tab:related_comparison} summarizes the comparison across methods. The key observation is that most existing methods perform corrections \textbf{under the premise of accepting unequal trajectory lengths}. This is akin to trying to calibrate a bent ruler---no matter how many correction factors are added, it is never as good as using a straight ruler in the first place. \method{} is that ``straight ruler'': it generates equal-length trajectory pairs at the inference framework level, mitigating bias from the generation side.

    \begin{table}[h]
        \begin{center}
            \resizebox{\textwidth}{!}{
                \begin{tabular}{lcccl}
                    \toprule
                    \textbf{Method} & \textbf{Modification Level} & \textbf{Equal-Length Guarantee} & \textbf{New Hyperparams} & \textbf{Core Limitation} \\
                    \midrule
                    Dr. GRPO & Loss function & No & 0 & Absolute gradient of long trajectories still biased \\
                    DAPO & Loss + reward & No & 1 & Introduces bias \\
                    GSPO & Loss function & No & 0 & Further amplifies length bias \\
                    LUSPO & Loss function & No & 0 & Absolute gradient of long trajectories still biased \\
                    DLER & Loss + sampling & No & 2 & Truncation may discard reasoning steps \\
                    GR3 & Loss function & No & 1 & Introduces bias \\
                    \midrule
                    \method{} & Inference framework & \textbf{Yes} & 0 & Requires validation on more scenarios \\
                    \bottomrule
                \end{tabular}
            }
        \end{center}
        \caption{Comparison of existing length bias mitigation methods. \method{} is the only method that guarantees equal length at the inference framework level and introduces no new hyperparameters.}\label{tab:related_comparison}
    \end{table}

    \section{Method}

    \subsection{Preliminaries: GRPO}

    Given a question $q$, the policy $\pi_\theta$ independently samples $G$ trajectories $\{o_1, \ldots, o_G\}$ (e.g., $G=16$). Each trajectory $o_i$ is scored by a verifiable reward function $R$ to obtain $r_i \in \{0, 1\}$. GRPO computes the advantage via group-level normalization:
    \begin{equation}
        \hat{A}_i = \frac{r_i - \text{mean}(\{r_j\}_{j=1}^G)}{\text{std}(\{r_j\}_{j=1}^G)}
        \label{eq:grpo_advantage}
    \end{equation}

    The policy gradient loss of GRPO is:
    \begin{equation}
        \mathcal{L}_{\text{GRPO}} = -\frac{1}{G}\sum_{i=1}^{G} \frac{1}{|o_i|}\sum_{t=1}^{|o_i|} \left[\min\left(\rho_{i,t}\hat{A}_i, \text{clip}(\rho_{i,t}, 1-\epsilon, 1+\epsilon)\hat{A}_i\right)\right]
        \label{eq:grpo_loss}
    \end{equation}
    where $\rho_{i,t} = \frac{\pi_\theta(o_{i,t}|q, o_{i,<t})}{\pi_{\text{old}}(o_{i,t}|q, o_{i,<t})}$ is the token-level importance ratio and $|o_i|$ is the length of trajectory $i$.

    \textbf{The source of length bias.} The $\frac{1}{|o_i|}$ term in Equation~\ref{eq:grpo_loss} is the direct source of length bias. Consider two trajectories $o_a$ (length 50, reward 1) and $o_b$ (length 500, reward 0): the gradient magnitude per token in $o_a$ is 10$\times$ that of $o_b$. Even if, as suggested by Dr.~GRPO~\citep{liu2025understanding}, $\frac{1}{|o_i|}$ is removed, the sum of gradients across all 500 tokens in the long trajectory $o_b$ still has a far larger absolute value than the sum across 50 tokens in $o_a$, biasing the optimization toward long trajectories.

    This reveals the \textbf{length bias dilemma}: as long as trajectories within a group differ in length, dividing by $|o_i|$ biases toward short responses (larger per-token gradients), while not dividing biases toward long responses (larger gradient sums)---the two normalization schemes each have their own bias direction, with no ``correct'' middle ground.

    \subsection{Equal-Length Trajectory Generation Framework}

    Our proposed \method{} generates equal-length trajectory pairs by modifying the inference process itself, eliminating the source of length bias.

    \subsubsection{Core Mechanism}

    \textbf{Parallel dual-track generation.} For each question $q$, we initialize two trajectories generated in parallel with $q$ as the shared prefix, decoding synchronously token by token:
    \begin{equation}
        a_t \sim \pi_\theta(\cdot | q, a_{<t}), \quad b_t \sim \pi_\theta(\cdot | q, b_{<t})
    \end{equation}
    The two trajectories are decoded in parallel within the same batch, sharing the KV cache for the question prefix, so the overhead compared to single-trajectory generation is negligible.

    \textbf{Equal-length segment extraction.} The two trajectories are generated synchronously. When one trajectory reaches the end-of-sequence token (EOS) first, equal-length segments are extracted from both trajectories up to that point. The trajectory that terminates first is denoted $A_i$, and the corresponding equal-length portion from the other trajectory is denoted $B_i$.
    Move the stopped trajectories out of the inference framework, and use the unstopped trajectories as prefixes to launch 2 new continuations.
    When the next trajectory reaches EOS, equal-length segments $A_{i+1}$ and $B_{i+1}$ are extracted again, and this cycle repeats until both trajectories have terminated. Using the computational budget for generating two trajectories, we effectively obtain many more trajectory pairs.

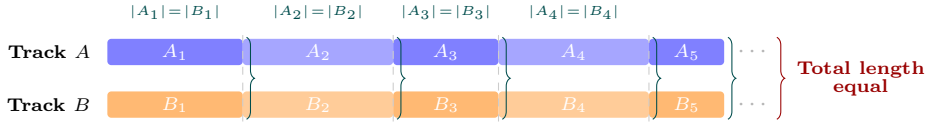
\begin{figure}[h]
        \centering
        \begin{tikzpicture}[
            lbl/.style={font=\scriptsize},
            brace/.style={decorate, decoration={brace, amplitude=3pt}},
        ]
        \def\xstart{0.0}
        \def\bh{0.35} 

        \node[font=\scriptsize\bfseries, anchor=east] at (\xstart-0.1, \bh/2) {Track $A$};

        \fill[blue!50, rounded corners=1.5pt] (\xstart, 0) rectangle (1.8, \bh);
        \node[font=\scriptsize\bfseries, white] at (0.9, \bh/2) {$A_1$};

        \fill[blue!35, rounded corners=1.5pt] (1.8, 0) rectangle (3.8, \bh);
        \node[font=\scriptsize\bfseries, white] at (2.8, \bh/2) {$A_2$};

        \fill[blue!50, rounded corners=1.5pt] (3.8, 0) rectangle (5.2, \bh);
        \node[font=\scriptsize\bfseries, white] at (4.5, \bh/2) {$A_3$};

        \fill[blue!35, rounded corners=1.5pt] (5.2, 0) rectangle (7.2, \bh);
        \node[font=\scriptsize\bfseries, white] at (6.2, \bh/2) {$A_4$};

        \fill[blue!50, rounded corners=1.5pt] (7.2, 0) rectangle (8.2, \bh);
        \node[font=\scriptsize\bfseries, white] at (7.7, \bh/2) {$A_5$};

        \node[font=\small, gray!60] at (8.6, \bh/2) {$\cdots$};

        \def\yb{-0.7}
        \node[font=\scriptsize\bfseries, anchor=east] at (\xstart-0.1, \yb+\bh/2) {Track $B$};

        \fill[orange!55, rounded corners=1.5pt] (\xstart, \yb) rectangle (1.8, \yb+\bh);
        \node[font=\scriptsize\bfseries, white] at (0.9, \yb+\bh/2) {$B_1$};

        \fill[orange!40, rounded corners=1.5pt] (1.8, \yb) rectangle (3.8, \yb+\bh);
        \node[font=\scriptsize\bfseries, white] at (2.8, \yb+\bh/2) {$B_2$};

        \fill[orange!55, rounded corners=1.5pt] (3.8, \yb) rectangle (5.2, \yb+\bh);
        \node[font=\scriptsize\bfseries, white] at (4.5, \yb+\bh/2) {$B_3$};

        \fill[orange!40, rounded corners=1.5pt] (5.2, \yb) rectangle (7.2, \yb+\bh);
        \node[font=\scriptsize\bfseries, white] at (6.2, \yb+\bh/2) {$B_4$};

        \fill[orange!55, rounded corners=1.5pt] (7.2, \yb) rectangle (8.2, \yb+\bh);
        \node[font=\scriptsize\bfseries, white] at (7.7, \yb+\bh/2) {$B_5$};

        \node[font=\small, gray!60] at (8.6, \yb+\bh/2) {$\cdots$};

        \draw[brace, teal!60!black] (1.85, \bh+0.02) -- (1.85, \yb-0.02);
        \draw[brace, teal!60!black] (3.85, \bh+0.02) -- (3.85, \yb-0.02);
        \draw[brace, teal!60!black] (5.25, \bh+0.02) -- (5.25, \yb-0.02);
        \draw[brace, teal!60!black] (7.25, \bh+0.02) -- (7.25, \yb-0.02);
        \draw[brace, teal!60!black] (8.25, \bh+0.02) -- (8.25, \yb-0.02);

        \draw[dashed, gray!40, line width=0.4pt] (1.8, \bh+0.05) -- (1.8, \yb-0.05);
        \draw[dashed, gray!40, line width=0.4pt] (3.8, \bh+0.05) -- (3.8, \yb-0.05);
        \draw[dashed, gray!40, line width=0.4pt] (5.2, \bh+0.05) -- (5.2, \yb-0.05);
        \draw[dashed, gray!40, line width=0.4pt] (7.2, \bh+0.05) -- (7.2, \yb-0.05);

        \node[font=\tiny, teal!60!black] at (0.9, \bh+0.35) {$|A_1|\!=\!|B_1|$};
        \node[font=\tiny, teal!60!black] at (2.8, \bh+0.35) {$|A_2|\!=\!|B_2|$};
        \node[font=\tiny, teal!60!black] at (4.5, \bh+0.35) {$|A_3|\!=\!|B_3|$};
        \node[font=\tiny, teal!60!black] at (6.2, \bh+0.35) {$|A_4|\!=\!|B_4|$};

        \draw[brace, red!60!black] (8.9, \bh+0.02) -- (8.9, \yb-0.02)
            node[midway, right=4pt, font=\scriptsize\bfseries, red!60!black, align=center] {Total length\\[-1pt]equal};

        \end{tikzpicture}
        \caption{Illustration of equal-length trajectory generation. Two tracks $A$ and $B$ sequentially produce multiple equal-length segment pairs $(A_i, B_i)$, each strictly equal in length ($|A_i|=|B_i|$).}
        \label{fig:method_detail}
    \end{figure}

    Figure~\ref{fig:method_detail} provides a concrete example.

    \subsubsection{Formal Description}

    The group size is set to $G$, and each group is split into subgroups of size 2. Each subgroup's generation process produces $N$ equal-length segment pairs $\{(A_i, B_i)\}_{i=1}^{N}$. The core invariant is the equal-length constraint:
    \begin{equation}
        |A_i| = |B_i|, \quad \forall i \in \{1, \ldots, N\}
        \label{eq:equal_length}
    \end{equation}

    \subsection{Reinforcement Learning with Equal-Length Trajectory Pairs}

    To avoid notational conflict with the advantage function $\hat{A}$, we henceforth denote each equal-length segment pair $(A_i, B_i)$ as $(o_i^+, o_i^-)$. Due to the prefix inheritance mechanism, the conditional input for each pair varies with the pair index: the first pair takes the original question $q$ as input, while the $i$-th pair ($i \geq 2$) uses $x_i = q \oplus \text{prefix}_i$, where $\text{prefix}_i$ is the inherited prefix sequence. Accordingly, the conditional input for each pair is uniformly denoted as $x_i$ ($x_1 = q$).

    \textbf{Reward computation.} For each equal-length segment pair, we use a verifiable reward $R$ (e.g., answer matching, rubric-based scoring) to evaluate both segments separately. Specifically, $o_i^+$ is scored directly by $R$; the reward for $o_i^-$ is taken as the maximum reward among all subsequent extension trajectories---if at least one correct trajectory can be generated from $o_i^-$, then $o_i^-$ itself constitutes a valid reasoning path. This design addresses a key deficiency of standard GRPO: in the standard approach, if a trajectory consists of a correct intermediate segment followed by an incorrect continuation, the entire trajectory is judged incorrect, and the correct intermediate segment also receives negative gradients, leading to erroneous updates. Our per-segment evaluation mechanism effectively avoids this issue. We conduct an ablation study comparing the maximum with the mean.

    \textbf{Skip rule.} When both segments in a pair receive the same reward, the pair is excluded from gradient updates. This mechanism naturally avoids the entropy collapse and wasted update problems analyzed in Section~\ref{sec:theory}.

    \textbf{Low-cost step-level rewards.}
    Our method effectively splits a single trajectory into multiple segments, each with a distinct reward value, indirectly upgrading outcome-level rewards to step-level rewards.

    Given the set of equal-length segment pairs $\mathcal{P} = \{(x_i, o_i^+, o_i^-)\}_{i=1}^{N}$, each pair constitutes an independent group (group size $G'=2$), and the advantage is computed separately for each group. Let the rewards be $r_i^+$ and $r_i^-$, and the advantages be $\hat{A}_i^+$ and $\hat{A}_i^-$.

    The policy gradient loss for each pair simplifies to:
    \begin{equation}
        \mathcal{L}_{\method{}} = -\frac{1}{2}\sum_{s \in \{+,-\}} \frac{1}{|o_i^s|}\sum_{t=1}^{|o_i^s|} \min\left(\rho_{i,t}^s \hat{A}_i^s, \text{clip}(\rho_{i,t}^s, 1-\epsilon, 1+\epsilon)\hat{A}_i^s\right)
        \label{eq:eqlen_loss}
    \end{equation}
    where $\rho_{i,t}^s = \frac{\pi_\theta(o_{i,t}^s | x_i, o_{i,<t}^s)}{\pi_{\text{old}}(o_{i,t}^s | x_i, o_{i,<t}^s)}$ is the token-level importance ratio, and the probability for each segment is computed conditioned on its respective input $x_i$.

    For the $G/2$ subgroups of a given input $x$ (the original group size $G$ split into $G/2$ subgroups of size 2), the $k$-th subgroup produces $N_k$ equal-length segment pairs. The total objective function is:
    \begin{equation}
        \mathcal{L}_{\text{total}} = \frac{1}{G/2}\sum_{k=1}^{G/2} \frac{1}{N_k}\sum_{i=1}^{N_k} \mathcal{L}_{\method{}}^{(k,i)}
        \label{eq:total_loss}
    \end{equation}
    where $\mathcal{L}_{\method{}}^{(k,i)}$ is the loss for the $i$-th segment pair in the $k$-th subgroup (Equation~\ref{eq:eqlen_loss}).

    The key property of Equation~\ref{eq:eqlen_loss}: since $|o_i^+| = |o_i^-|$, the two segments within each pair receive exactly the same scaling regardless of whether the $\frac{1}{|o_i^s|}$ normalization term is included. This means that \textbf{length is no longer a variable influencing the relative optimization direction within a group}---precisely the goal that Dr.~GRPO, GSPO, LUSPO, and other methods attempt but fail to fully achieve.

    \subsection{Advantages of Prefix Inheritance}

    \textbf{Advantages of prefix inheritance.} The prefix inheritance mechanism is conceptually consistent with prefix tuning~\citep{li-liang-2021-prefix}, chain-of-thought prompting~\citep{NEURIPS2022_9d560961}, and prefix-conditioned generation in rejection sampling. Prefix inheritance is equivalent to masking the prefix so that it does not participate in gradient updates, indirectly upgrading outcome-level rewards to step-level rewards.

    \section{Theoretical Analysis}
    \label{sec:theory}

    \subsection{Failure Modes from Length Inconsistency}

    Existing research generally holds that the primary harm of length inconsistency in GRPO is causing model responses to become progressively longer or shorter (length bias), as rigorously proven by Dr.~GRPO~\citep{liu2025understanding}. However, the harms of length inconsistency extend far beyond this. This section proves through constructive counterexamples that GRPO \textbf{necessarily} produces two additional failure modes under certain conditions (complete proofs in Appendix~\ref{app:proofs}).

    The following analysis is based on the GRPO loss (Equation~\ref{eq:grpo_loss}), with $G=2$, binary rewards $r \in \{0, 1\}$, and the PPO clipping term omitted (irrelevant to this paper). The gradient contribution of trajectory $i$ is $g_i = -\frac{\hat{A}_i}{|o_i|}\sum_{t=1}^{|o_i|} \nabla_\theta \log \pi_\theta(o_{i,t}|q, o_{i,<t})$.

    \textbf{Proposition 1 (Existence of entropy collapse).} \textit{There exists a question $q$ and a policy $\pi_\theta$ such that a single GRPO update step strictly decreases the prefix generation probability of a certain correct solution.} Intuition: when a short correct trajectory is paired with a long incorrect trajectory (containing a correct prefix), $\hat{A}_2 = -1$ causes \textbf{all} tokens of the long trajectory (including the correct prefix) to receive negative gradients. This event has nonzero probability, and its cumulative effect systematically suppresses long-solution patterns.

    \textbf{Proposition 2 (Existence of learning tax).} \textit{There exists a prefix $P$ and a policy $\pi_\theta$ such that the expected gradient of $P$ during GRPO training is zero, but the gradient variance is strictly positive.} Intuition: when the prefix $P$ has no predictive power for the final answer ($p=0.5$), the tokens of $P$ alternately receive positive and negative gradients, yielding $\mathbb{E}[g_t] = \mathbf{0}$ but $\text{Var}(g_t) = \|v_t\|^2/L^2 > 0$, producing $O(\sqrt{T})$ parameter drift that constitutes pure noise overhead.

    \textbf{How \method{} avoids these failure modes.} \method{} simultaneously eliminates the triggering conditions for both failure modes: (1) inherited prefixes are masked and excluded from gradient computation, so the erroneous penalization in Proposition~1 does not occur; (2) each equal-length segment pair is scored independently, and segments contain no reward-irrelevant shared prefixes, so the gradient oscillation in Proposition~2 does not arise.

    \section{Experiments}

    We conduct experiments under a compute-matched setting (equal total GPU hours) on Qwen3-32B (dense) and Qwen3-Next (MoE), evaluating on AIME25, HMMT25, and LiveCodeBench against five baselines: GRPO, Dr.~GRPO, DAPO, GSPO, and LUSPO. All experiments use 5 random seeds, reporting means $\pm$ 95\% bootstrap confidence intervals. Full experimental setup details are provided in Appendix~\ref{app:exp_setup}.

    \subsection{Main Results}

    \begin{table}[h]
        \begin{center}
            \small
            \setlength{\tabcolsep}{4pt}
            \renewcommand{\arraystretch}{1.15}
            \resizebox{\textwidth}{!}{
            \begin{tabular}{lcccccc}
                \toprule
                \multirow{2}{*}{\textbf{Method}} &
                \multicolumn{3}{c|}{\textbf{Qwen3-32B Acc avg@32 (\%)}} &
                \multicolumn{3}{c}{\textbf{Qwen3-Next Acc avg@32 (\%)}} \\
                \cmidrule(lr){2-4}\cmidrule(lr){5-7}
                & \textbf{AIME25} & \textbf{LiveCodeBench} & \textbf{HMMT25}
                & \textbf{AIME25} & \textbf{LiveCodeBench} & \textbf{HMMT25} \\
                \midrule
                Base & 72.9 & 60.6 & 51.5 & 87.8 & 68.7 & 73.9 \\
                GRPO & 75.8$\pm$1.0 & 61.2$\pm$0.8 & 54.6$\pm$1.1 & 88.0$\pm$1.3 & 69.2$\pm$0.7 & 74.8$\pm$1.2 \\
                Dr. GRPO & 77.4$\pm$0.9 & 62.5$\pm$0.9 & 56.1$\pm$1.0 & 89.1$\pm$1.2 & 70.0$\pm$0.8 & 76.2$\pm$1.1 \\
                DAPO & 78.2$\pm$1.0 & 63.1$\pm$0.8 & 57.8$\pm$1.1 & 89.5$\pm$1.3 & 70.4$\pm$0.9 & 76.8$\pm$1.2 \\
                GSPO & 76.6$\pm$1.1 & 61.8$\pm$0.9 & 55.4$\pm$1.2 & 88.4$\pm$1.4 & 69.6$\pm$0.8 & 75.4$\pm$1.4 \\
                LUSPO & 78.8$\pm$0.9 & 63.4$\pm$0.8 & 58.2$\pm$1.0 & 89.8$\pm$1.2 & 70.8$\pm$0.7 & 77.1$\pm$1.1 \\
                \method{}-GRPO & \textbf{84.2$\pm$1.0} & \textbf{68.5$\pm$0.9} & \textbf{64.0$\pm$1.0} & \textbf{92.6$\pm$1.1} & \textbf{74.2$\pm$1.2} & \textbf{82.8$\pm$0.9} \\
                \bottomrule
            \end{tabular}
            }
        \end{center}
        \caption{Main results under compute-matched setting (equal total GPU hours; 5-seed mean $\pm$ 95\% bootstrap CI, Acc avg@32, \%). \method{}-GRPO outperforms all tested baselines across three benchmarks and two architectures; all confidence intervals are non-overlapping with the strongest baseline, indicating statistically significant differences.}\label{tab:main}
    \end{table}

    Table~\ref{tab:main} presents the main experimental results. \method{} achieves the best performance across all benchmarks and model architectures:

    \textbf{Accuracy.} On Qwen3-32B, \method{} achieves 84.2\% on AIME25 (surpassing the strongest baseline LUSPO by 5.4 percentage points), 64.0\% on HMMT25 (+5.8\%), and 68.5\% on LiveCodeBench (+5.1\%). Consistent large improvements are also observed on Qwen3-Next (AIME25: +2.8\%, HMMT25: +5.7\%, LiveCodeBench: +3.4\%), validating the method's generalization across different model architectures (dense and MoE).

    \textbf{Advantage on long-chain reasoning.} \method{}'s advantage is most pronounced on HMMT25, which requires long-chain reasoning (Qwen3-32B: +5.8\%, Qwen3-Next: +5.7\%). This is because longer reasoning chains exhibit greater trajectory length variation, making length bias more harmful, and \method{}'s equal-length mechanism precisely eliminates this bottleneck.

    \textbf{Training stability.} We observe that \method{}'s training curves rise monotonically throughout the training process, without the sharp oscillations commonly seen in standard GRPO and GSPO. This is attributed to the cleaner gradient signals provided by equal-length trajectory pairs, which reduce the gradient noise introduced by length differences.

    \subsection{Generation Efficiency Analysis}
    \label{sec:efficiency_exp}

    \begin{table}[h]
        \begin{center}
            \resizebox{\columnwidth}{!}{
            \begin{tabular}{lccccc}
                \toprule
                \textbf{Method} & \textbf{Parallel Tracks} & \textbf{Eff. Samples} & \textbf{GPU Time (h)} & \textbf{Samples/GPU-h} & \textbf{Equal-Len.} \\
                \midrule
                GRPO ($G$=16) & 16 & 8,520 & 48.0 & 178 & No \\
                \method{}-GRPO & 16 & 51,456 & 48.0 & \textbf{1,072} & \textbf{Yes} \\
                \bottomrule
            \end{tabular}
            }
        \end{center}
        \caption{Generation efficiency comparison (Qwen3-32B). Effective samples denotes the number of training samples receiving non-zero gradient updates: in GRPO, this is the number of trajectories with non-zero advantage; in \method{}, this is the number of equal-length segment pairs $\times$ 2 (each pair produces 2 samples). \method{} uses 16 parallel tracks (8 subgroups of 2), but through prefix inheritance produces approximately 6 equal-length pairs per generation budget, yielding roughly 6$\times$ the efficiency of GRPO.}\label{tab:efficiency}
    \end{table}

    Table~\ref{tab:efficiency} shows that \method{}-GRPO achieves approximately 6$\times$ the sample efficiency of GRPO.

    \subsection{Ablation Study}

    \begin{table}[t]
        \begin{center}
            \begin{tabular}{lcc}
                \toprule
                \textbf{Variant} & \textbf{AIME25} & \textbf{HMMT25} \\
                \midrule
                \method{}-GRPO (full) & \textbf{84.2$\pm$1.0} & \textbf{64.0$\pm$1.0} \\
                Single-pair mode & 77.5$\pm$0.8 & 57.6$\pm$1.3 \\
                Prefix in gradient & 76.8$\pm$1.1 & 57.2$\pm$1.2 \\
                Mean reward & 80.6$\pm$1.0 & 60.4$\pm$1.1 \\
                \bottomrule
            \end{tabular}
        \end{center}
        \caption{Ablation study results (Qwen3-32B, 5-seed mean $\pm$ 95\% bootstrap CI, Acc avg@32, \%).}\label{tab:ablation}
    \end{table}

    Table~\ref{tab:ablation} validates the contribution of each component: (1)~\textbf{Single-pair mode} (generation stops when one trajectory terminates without prefix inheritance for continued generation, collecting only one equal-length pair per question) causes a drop on AIME25, as the number of available training pairs is substantially reduced and subsequent steps cannot benefit from contrastive learning; (2)~\textbf{Prefix in gradient} (the inherited prefix is not masked and participates in gradient updates) causes a drop, as prefix tokens receive reward signals mismatched with their generation context, introducing noisy gradients that interfere with learned representations; (3)~\textbf{Mean reward} (replacing the maximum with the mean over subsequent extension trajectories for the reward of $o_i^-$) causes a drop, as the mean is dragged down by incorrect extensions, diluting the positive signal from correct intermediate segments.

    \subsection{Combination with Existing Methods}

    \method{} can be orthogonally combined with existing algorithms. On Qwen3-32B (AIME25 Acc avg@32): \method{}-Dr.~GRPO reaches 85.1\% (+0.9\%), and \method{}-DAPO reaches \textbf{86.4\%} (+2.2\%), both exceeding \method{}-GRPO's standalone 84.2\%. This demonstrates that equal-length generation mitigates length bias at the generation level, while loss function improvements enhance gradient signal \textit{quality}---the two are complementary.

    \section{Discussion and Limitations}

    \textbf{Why is equal length so important?} The essence of length bias is the coupling between length and reward signals in the loss function. \method{} decouples these two signals by enforcing equal length, so that the model can only obtain optimization reward by improving answer quality.

    \textbf{Limitations and future directions.} (1)~Our experiments focus on verifiable reward scenarios in mathematical and code reasoning; effectiveness in open-domain generation with reward model-based scoring remains to be validated.

    \section{Conclusion}

    This paper proposes \method{}, a method that mitigates length bias in group-relative reinforcement learning at the inference framework level. The core mechanisms---synchronous dual-track generation and prefix inheritance---naturally produce equal-length trajectory pairs, preventing length from interfering with gradient computation. Experiments demonstrate that \method{} outperforms all tested baselines on both Qwen3-32B and Qwen3-Next architectures, and can be orthogonally combined with existing loss function improvements. We believe the paradigm introduced by \method{} also offers insights for addressing other systematic biases in reinforcement learning training, such as difficulty bias and style bias.

    \section*{Reproducibility Statement}
    Code will be released.

    \appendix

    \section{Experimental Setup}
    \label{app:exp_setup}

    \textbf{Overall setting.} We compare \method{} against baseline methods under a compute-matched setting (equal total GPU hours). Specifically, using \method{}'s training GPU hours as the reference, each baseline adjusts its number of training steps to consume the same total GPU hours, ensuring a fair comparison. The evaluation metric is Acc avg@32 (average accuracy over 32 independent samples).

    \textbf{Base models.} We use Qwen3-32B (dense architecture) and Qwen3-Next-80B-A3B-Thinking (MoE architecture, 3B active parameters), covering both dense and MoE mainstream architectures.

    \textbf{Training data.} We use a mixed subset of DeepMath-103K~\citep{he2025deepmath103klargescalechallengingdecontaminated} (decontaminated mathematical reasoning dataset) and OpenCodeReasoning~\citep{ahmad2025opencodereasoning} (decontaminated coding reasoning dataset). The reward function is exact answer matching ($r=1$ if the final answer is correct, $r=0$ otherwise), requiring no trained reward model.

    \textbf{Evaluation benchmarks.} We cover three types of reasoning tasks: AIME25~\citep{maa_aime2025} (competition mathematics), HMMT25~\citep{balunovic2026matharenaevaluatingllmsuncontaminated} (long-chain mathematical reasoning), and LiveCodeBench v6~\citep{jain2024livecodebenchholisticcontaminationfree} (code reasoning).

    \textbf{Baselines.} We select 5 representative baselines covering the main technical approaches to length bias mitigation: \textbf{GRPO}~\citep{shao2024deepseekmathpushinglimitsmathematical} (standard implementation, $G=16$), \textbf{Dr.~GRPO}~\citep{liu2025understanding} (removing length and standard deviation normalization), \textbf{DAPO}~\citep{yu2026dapo} (token-level loss + overlong reward shaping), \textbf{GSPO}~\citep{zheng2025groupsequencepolicyoptimization} (sequence-level optimization), and \textbf{LUSPO}~\citep{liu2026lengthunbiasedsequencepolicyoptimization} (length-scaled GSPO correction).

    \textbf{Implementation details.} All methods use the same training and inference hyperparameters. All experiments are run with 5 random seeds, reporting means $\pm$ 95\% bootstrap confidence intervals.

    \section{Hyperparameter Configuration}
    \label{app:impl_details}

    \textbf{Training hyperparameters.} All methods use the same configuration: AdamW optimizer (20-step warmup), learning rate $1\times10^{-6}$ (Qwen3-32B) / $5\times10^{-7}$ (Qwen3-Next), KL penalty $\beta_{\text{kl}}=0.002$, maximum sequence length 32768, prompt batch size 64. Qwen3-32B is trained on 8$\times$A100 80GB, and Qwen3-Next on 16$\times$A100 80GB.

    \textbf{Inference settings.} Temperature\,=\,0.6, TopP\,=\,0.95, TopK\,=\,20, MinP\,=\,0. All methods are compared under the same decoding settings, ensuring that differences stem solely from the training method. All evaluations use 32 independent samples to compute avg@32.

    \textbf{Seeds and statistics.} All experiments are run with 5 random seeds, reporting means $\pm$ 95\% bootstrap confidence intervals.

    \textbf{\method{}-specific configuration.} Sampling group size $G=16$ (split into 8 subgroups of size 2); each question generates one dual-track stream, with each subgroup producing approximately 6 equal-length pairs on average.

    \section{Complete Proofs of Propositions 1 and 2}
    \label{app:proofs}

    \subsection{Proof of Proposition 1 (Existence of Entropy Collapse)}

    \textbf{Proposition 1.} \textit{There exists a question $q$ and a policy $\pi_\theta$ such that a single GRPO ($G=2$) update step strictly decreases the prefix generation probability of a certain correct solution.}

    \textit{Proof.} We adopt the equal-length segment notation from \method{}. Let two trajectories be generated in parallel, producing equal-length segment pairs $(A_i, B_i)$ where $|A_i| = |B_i|$ ($A_i$ denotes the segment that first reaches EOS, $B_i$ denotes the corresponding equal-length counterpart). Let question $q$ admit three possible responses:
    \begin{itemize}
        \item $A_1$: the first segment directly yields the correct answer (first to reach EOS), $r=1$, length $|A_1|$
        \item $B_1 + A_2$: $B_1$ is equal-length to $A_1$ ($|B_1|=|A_1|$) and serves as a valid reasoning prefix; $A_2$ is the subsequent segment but ultimately gives an incorrect answer, $r=0$, total length $|B_1|+|A_2|$
        \item $B_1 + B_2 + A_3$: shares the same prefix $B_1$ ($|B_1|=|A_1|$), $B_2$ is equal-length to $A_2$ ($|B_2|=|A_2|$), $A_3$ is the third segment that gives the correct answer, $r=1$, total length $|B_1|+|B_2|+|A_3|$
    \end{itemize}
    Here $B_1$ is a valid reasoning prefix (e.g., a problem modeling step such as ``let $x$ be...''), shared by both the correct response $B_1+B_2+A_3$ and the incorrect response $B_1+A_2$.

    \textbf{How GRPO handles this.} Standard GRPO treats entire trajectories as units, sampling $G=2$ trajectories and computing advantages from sequence-level rewards. Consider the following sampling outcome (this event has nonzero probability under $\pi_\theta$): $o_1 = A_1$ (correct, $r_1=1$) and $o_2 = B_1 + A_2$ (incorrect, $r_2=0$). The two trajectories have unequal lengths: $|o_1| = |A_1|$, $|o_2| = |B_1| + |A_2| = |A_1| + |A_2|$.

    From $r_1=1, r_2=0$, Equation~\ref{eq:grpo_advantage} yields $\hat{A}_2 = -1$. GRPO \textbf{uniformly broadcasts} this advantage to every token of $o_2$. The gradient contribution from $o_2$ is:
    \begin{equation}
        g_2 = \frac{1}{|o_2|}\sum_{t=1}^{|o_2|} \nabla_\theta \log \pi_\theta(o_{2,t}|q, o_{2,<t})
    \end{equation}

    The critical issue: within $o_2 = B_1 + A_2$, the first $|B_1|=|A_1|$ tokens of prefix $B_1$ are entirely correct reasoning steps, yet because $\hat{A}_2=-1$ is broadcast to the entire trajectory, \textbf{every token} in $B_1$ receives negative gradients.

    The parameter update is $\theta' = \theta - \eta g_2$ ($\eta > 0$). For any token $b_t$ in $B_1$ ($t \leq |B_1|$), by the positive-definiteness of the Fisher information matrix:
    \begin{equation}
        \nabla_\theta \log \pi_\theta(b_t|\cdot)^\top g_2 \geq \frac{1}{|o_2|}\|\nabla_\theta \log \pi_\theta(b_t|\cdot)\|^2 > 0
    \end{equation}

    Therefore $\pi_{\theta'}(b_t|q, b_{<t}) < \pi_\theta(b_t|q, b_{<t})$: the generation probability of prefix $B_1$ is \textbf{strictly decreased}. When the pairing $(A_1,\; B_1+A_2)$ is repeatedly encountered during training, $B_1$ is penalized again and again, causing \textbf{all} solutions starting with $B_1$---including the correct solution $B_1+B_2+A_3$---to be systematically suppressed. The model gradually loses the ability to reach correct answers through long reasoning chains, and policy diversity declines---i.e., \textbf{entropy collapse}.

    \textbf{How \method{} avoids this.} \method{} does not operate on entire trajectories but instead independently evaluates each equal-length segment pair $(A_i, B_i)$. In the above construction, \method{} evaluates $(A_1, B_1)$ as the first pair and $(A_2, B_2)$ as the second pair separately. For the first pair, $A_1$ is correct, and $B_1$'s reward is taken as the maximum reward among all its subsequent extended trajectories, so $B_1$ is not erroneously penalized and receives the correct reward. When $B_1$ serves as an inherited prefix, it is masked and excluded from gradient computation. Therefore $B_1$ is never penalized due to $A_2$'s error, and the failure mode of Proposition~1 does not occur. $\square$

    \subsection{Proof of Proposition 2 (Existence of Learning Tax)}

    \textbf{Proposition 2.} \textit{There exists a shared prefix $B_1$ and a policy $\pi_\theta$ such that the expected gradient of $B_1$ during GRPO training is zero, but the gradient variance is strictly positive---i.e., $B_1$ receives no effective learning signal yet continuously consumes optimizer resources.}

    \textit{Proof.} We continue with the construction and equal-length segment notation from Proposition~1. Question $q$ has three responses: $A_1$ (correct), $B_1+A_2$ (incorrect), $B_1+B_2+A_3$ (correct), where $|A_i|=|B_i|$. Let the policy $\pi_\theta$ intermittently sample trajectories containing prefix $B_1$.

    In GRPO with $G=2$, $B_1$ alternately receives opposing gradient signals under different pairings (when $r=r'$, $\hat{A}=0$ and no update occurs; we ignore this case):
    \begin{enumerate}
        \item \textbf{$B_1$ is penalized}: the pair $(A_1,\; B_1+A_2)$ is sampled. The entire $B_1+A_2$ is incorrect ($r=0$), $A_1$ is correct ($r'=1$), so $\hat{A}=-1$ is broadcast to every token of $B_1+A_2$. The gradient contribution of each token $b_t$ in $B_1$ is $g_t^{(-)} = +\frac{1}{|B_1|+|A_2|}\nabla_\theta \log \pi_\theta(b_t|\cdot)$ (decreasing $B_1$'s probability).
        \item \textbf{$B_1$ is rewarded}: the pair $(B_1+A_2,\; B_1+B_2+A_3)$ is sampled. $B_1+B_2+A_3$ is correct ($r=1$), $B_1+A_2$ is incorrect ($r'=0$), so $\hat{A}=+1$ is broadcast to every token of $B_1+B_2+A_3$. The gradient contribution of each token $b_t$ in $B_1$ is $g_t^{(+)} = -\frac{1}{|B_1|+|B_2|+|A_3|}\nabla_\theta \log \pi_\theta(b_t|\cdot)$ (increasing $B_1$'s probability).
    \end{enumerate}

    Let $L^{(-)} = |B_1|+|A_2|$, $L^{(+)} = |B_1|+|B_2|+|A_3|$, and $v_t = \nabla_\theta \log \pi_\theta(b_t|\cdot)$. Let the two cases occur with probabilities $p$ and $1-p$ respectively. When $\frac{p}{L^{(-)}} = \frac{1-p}{L^{(+)}}$ (i.e., $p = \frac{L^{(-)}}{L^{(-)}+L^{(+)}}$):
    \begin{align}
        \mathbb{E}[g_t] &= p \cdot \frac{v_t}{L^{(-)}} + (1-p) \cdot \left(-\frac{v_t}{L^{(+)}}\right) = \mathbf{0}
    \end{align}

    Yet the gradient variance is strictly positive:
    \begin{align}
        \text{Var}(g_t) &= \mathbb{E}[\|g_t\|^2] = p \cdot \frac{\|v_t\|^2}{(L^{(-)})^2} + (1-p) \cdot \frac{\|v_t\|^2}{(L^{(+)})^2} > 0
    \end{align}

    Prefix $B_1$ is \textbf{repeatedly penalized and rewarded} during training: penalized when the pair $(A_1,\; B_1+A_2)$ is encountered, rewarded when the pair $(B_1+A_2,\; B_1+B_2+A_3)$ is encountered, yet the expected gradient is zero---$B_1$ \textbf{learns no effective signal}. However, the per-step gradient variance is strictly positive, causing parameters to oscillate between positive and negative directions. Over $T$ training steps, the standard deviation of cumulative parameter drift is $O(\sqrt{T})$ (random walk), producing ineffective parameter drift that consumes optimizer momentum and learning budget, constituting a \textbf{learning tax}.

    \textbf{How \method{} avoids this.} In \method{}, each equal-length segment pair $(A_i, B_i)$ is independently evaluated, and the inherited prefix $B_1$ is masked and excluded from gradient computation. In the second pair $(A_2, B_2)$, $A_2$ and $B_2$ are equal-length and independently scored, so sequence-level rewards are not back-propagated to irrelevant prefixes. Therefore $B_1$ never receives contradictory gradients, and the learning tax of Proposition~2 does not occur. $\square$

    \section{Derivation of \method{} Integrated into RLOO}
    \label{app:eqlen_rloo}

    This section derives the complete loss function when \method{}'s equal-length generation mechanism is integrated into the RLOO~\citep{kool2019buy} framework.

    \subsection{Standard RLOO}

    RLOO~\citep{kool2019buy} samples $G$ independent trajectories $\{o_1, \ldots, o_G\}$ for each question $x$ and computes the advantage using a leave-one-out mean as the baseline:
    \begin{equation}
        \hat{A}_i^{\text{RLOO}} = r_i - \frac{1}{G-1}\sum_{j \neq i} r_j
        \label{eq:rloo_adv}
    \end{equation}
    The corresponding token-level policy gradient loss is:
    \begin{equation}
        \mathcal{L}_{\text{RLOO}} = -\frac{1}{G}\sum_{i=1}^{G} \frac{1}{|o_i|}\sum_{t=1}^{|o_i|} \hat{A}_i^{\text{RLOO}} \log \pi_\theta(o_{i,t}|x, o_{i,<t})
        \label{eq:rloo_loss}
    \end{equation}

    The advantage of RLOO is that it does not require standard deviation normalization (cf. GRPO's Equation~\ref{eq:grpo_advantage}), but the $\frac{1}{|o_i|}$ term in Equation~\ref{eq:rloo_loss} still applies different scaling to trajectories of different lengths, so length bias persists.

    \subsection{Integrating Equal-Length Generation}

    The $G$ trajectories are split into $G/2$ subgroups of size 2. Each subgroup produces $N_k$ equal-length segment pairs $\{(o_{k,i}^+, o_{k,i}^-)\}_{i=1}^{N_k}$ through \method{}'s synchronous dual-track generation and prefix inheritance mechanism, satisfying $|o_{k,i}^+| = |o_{k,i}^-| = L_{k,i}$.

    Applying RLOO with $G'=2$ independently to each equal-length segment pair (Equation~\ref{eq:rloo_adv}), the advantage simplifies to:
    \begin{equation}
        \hat{A}_{k,i}^{+} = r_{k,i}^+ - r_{k,i}^-, \qquad \hat{A}_{k,i}^{-} = r_{k,i}^- - r_{k,i}^+
        \label{eq:eqlen_rloo_adv}
    \end{equation}

    When $r_{k,i}^+ = r_{k,i}^-$, $\hat{A}_{k,i}^{\pm} = 0$ and the gradient is naturally zero, which is consistent with \method{}'s skip rule---no additional mechanism is needed.

    \subsection{EqLen-RLOO Loss Function}

    Substituting Equation~\ref{eq:eqlen_rloo_adv} into the RLOO loss (Equation~\ref{eq:rloo_loss}) and expanding for each equal-length segment pair:
    \begin{equation}
        \mathcal{L}_{\text{EqLen-RLOO}}^{(k,i)} = -\frac{1}{2}\sum_{s \in \{+,-\}} \frac{1}{L_{k,i}}\sum_{t=1}^{L_{k,i}} \hat{A}_{k,i}^s \log \pi_\theta(o_{k,i,t}^s | x_{k,i}, o_{k,i,<t}^s)
        \label{eq:eqlen_rloo_pair}
    \end{equation}
    where $x_{k,i}$ is the conditional input for the $i$-th pair in the $k$-th subgroup (including the inherited prefix). Since $|o_{k,i}^+| = |o_{k,i}^-| = L_{k,i}$, $\frac{1}{L_{k,i}}$ applies \textbf{exactly the same} scaling to $o^+$ and $o^-$---length bias is eliminated.

    Aggregating across all subgroups and segment pairs, the total loss is:
    \begin{equation}
        \mathcal{L}_{\text{EqLen-RLOO}} = \frac{1}{G/2}\sum_{k=1}^{G/2} \frac{1}{N_k}\sum_{i=1}^{N_k} \mathcal{L}_{\text{EqLen-RLOO}}^{(k,i)}
        \label{eq:eqlen_rloo_total}
    \end{equation}

    \subsection{Relationship to EqLen-GRPO}

    Comparing Equation~\ref{eq:eqlen_rloo_pair} with the \method{}-GRPO loss (Equation~\ref{eq:eqlen_loss}), the differences are limited to:
    \begin{enumerate}
        \item \textbf{Advantage function}: RLOO uses $\hat{A}^s = r^s - r^{\bar{s}}$ (no standard deviation normalization), while GRPO uses $\hat{A}^s = (r^s - \bar{r})/\sigma_r$. Under binary rewards $r \in \{0,1\}$ with $r^+ \neq r^-$, both reduce to $\pm 1$, making them \textbf{exactly equivalent}.
        \item \textbf{PPO clipping}: \method{}-GRPO includes the clipping mechanism $\min(\rho_t \hat{A}, \text{clip}(\rho_t, 1{-}\epsilon, 1{+}\epsilon)\hat{A})$, while pure RLOO does not. In practice, clipping can be freely added for enhanced stability.
    \end{enumerate}

    Therefore, under binary rewards, \method{}-RLOO and \method{}-GRPO are identical in their advantage signals, with PPO clipping as an optional stability enhancement. The structure of the total loss (Equations~\ref{eq:eqlen_rloo_total} and~\ref{eq:total_loss}) is also identical, demonstrating that \method{}'s equal-length generation mechanism can serve as a plug-and-play module that seamlessly integrates into different policy optimization frameworks such as RLOO and GRPO.

\end{document}